\documentclass{article}

\usepackage[final]{neurips_2025}
\usepackage{algorithm}
\usepackage{algpseudocode}
\usepackage{multirow}
\usepackage[table]{xcolor}

\usepackage[utf8]{inputenc} 
\usepackage[T1]{fontenc}    
\usepackage[colorlinks=true, citecolor=blue, linkcolor=blue, urlcolor=blue]{hyperref}
\usepackage{url}            
\usepackage{booktabs}       
\usepackage{amsfonts}       
\usepackage{nicefrac}       
\usepackage{textcomp} 
\usepackage{microtype}      
\usepackage{xcolor}         
\usepackage{graphicx}
\usepackage{amsmath}
\usepackage{gensymb}
\usepackage{subcaption}
\usepackage{caption}

\makeatletter
\renewcommand{\paragraph}{%
  \@startsection{paragraph}{4}%
  {\z@}{0ex \@plus 0ex \@minus .2ex}{-0.5em}%
  {\normalfont\normalsize\bfseries}%
}

\title{DiffEye: Diffusion-Based Continuous Eye-Tracking Data Generation Conditioned on Natural Images}

\author{%
  Ozgur Kara\thanks{Equal contribution} \quad Harris Nisar\footnotemark[1] \quad James M. Rehg \\
  \texttt{\{ozgurk2, nisar2, jrehg\}@illinois.edu} \\
  University of Illinois Urbana-Champaign \\
}

\begin{document}

\maketitle

\begin{abstract}
Numerous models have been developed for scanpath and saliency prediction, which are typically trained on scanpaths, which model eye movement as a sequence of discrete fixation points connected by saccades, while the rich information contained in the raw trajectories is often discarded. Moreover, most existing approaches fail to capture the variability observed among human subjects viewing the same image. They generally predict a single scanpath of fixed, pre-defined length, which conflicts with the inherent diversity and stochastic nature of real-world visual attention.
To address these challenges, we propose \texttt{DiffEye}, a diffusion-based training framework designed to model continuous and diverse eye movement trajectories during free viewing of natural images. Our method builds on a diffusion model conditioned on visual stimuli and introduces a novel component, namely \textit{Corresponding Positional Embedding (CPE)}, which aligns spatial gaze information with the patch-based semantic features of the visual input. By leveraging raw eye-tracking trajectories rather than relying on scanpaths, \texttt{DiffEye} captures the inherent variability in human gaze behavior and generates high-quality, realistic eye movement patterns, despite being trained on a comparatively small dataset. The generated trajectories can also be converted into scanpaths and saliency maps, resulting in outputs that more accurately reflect the distribution of human visual attention.
\texttt{DiffEye} is the first method to tackle this task on natural images using a diffusion model while fully leveraging the richness of raw eye-tracking data. Our extensive evaluation shows that \texttt{DiffEye} not only achieves state-of-the-art performance in scanpath generation but also enables, for the first time, the generation of continuous eye movement trajectories. Project webpage: \href{https://diff-eye.github.io/}{https://diff-eye.github.io/}
\end{abstract}

\section{Introduction} \label{sec:intro}
Humans obtain visual information via a sequence of eye movements that direct the fovea to analyze important elements of the visual scene. The resulting pattern of fixations and saccades\footnote{Fixations occur when the eye remains relatively still and attention is focused on a specific location. 
Saccades are rapid eye movements between fixations that enable quick shifts of focus between regions of interest.} is dependent upon the task at hand (e.g., free exploration, visual search, etc.) and determines the visual elements that receive our attention~\cite{yarbus2013eye, Rothkopf2016}. 
The gold standard for quantifying visual attention is eye-tracking, which records a subject’s eye movements as they view an image (stimulus) on a monitor. Eye tracking produces a trajectory of eye movements, consisting of a sequence of $(x,y)$ pairs, which is processed to identify the fixations and saccades. Eye tracking provides valuable insights into the properties of the visual scene that drive human attention, and is widely-used for applications in virtual reality \cite{Clay2019, Razzak2025, Ren2024}, foveated rendering \cite{Kim2024, Liu2025}, and advertising \cite{Liu2020, Zhang2018}. Eye tracking is also widely used in psychology to identify differences in how visual information is processed by individuals~\cite{holmqvist2011eye}. For example, eye tracking has been used to study social cognition in conditions such as autism by identifying differences in how social cues are processed~\cite{nayar22,belen23} (e.g., individuals with autism have been shown to exhibit atypical patterns of attention to faces~\cite{McPartland2010,chawarska10}). Although eye-tracking studies provide high-quality data, they are costly and time-consuming to conduct, and require significant expertise for proper experimental design \cite{holmqvist2011eye}. As a result, there has been growing interest in developing computational models that can predict visual attention to images.



\begin{figure}
    \centering
    \includegraphics[width=.85\linewidth]{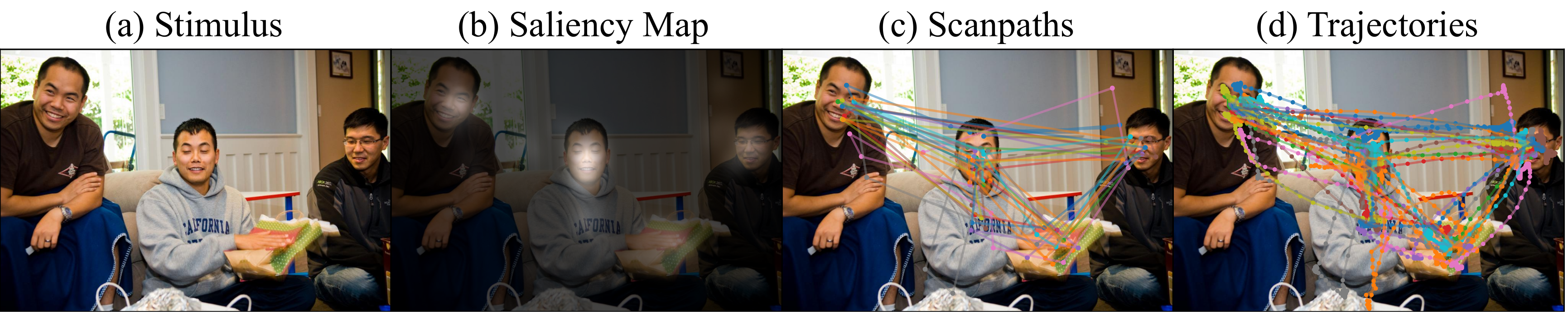}
\caption{\textbf{\textit{Comparison of different eye-tracking data types.}} 
(a) Original visual stimulus. 
(b) Saliency maps highlight regions of interest but do not capture the temporal dynamics of human attention. 
(c) Scanpaths offer a compressed representation of eye movement trajectories. 
(d) Full eye movement trajectories, recorded via eye trackers, provide detailed insights into attention dynamics. 
This example is from the MIT1003 dataset~\cite{Judd2009}; each color represents a different subject, emphasizing the importance of modeling inter-subject variability.
    } 
    \label{fig:heatmap-scanpath-cont}
\end{figure}

Modeling human visual attention has traditionally focused on saliency prediction~\cite{bylinskii2016should, borji2019saliency}. In this task, given an image, the goal is to produce a heat map, known as a saliency map, where high values correspond to regions that are more likely to receive fixations. However, these spatial summaries neglect the temporal structure of eye movements, which is critical for applications in virtual reality and psychology research. For example, recent work in developmental science has shown that children with and without autism demonstrate different spatio-temporal gaze behaviors~\cite{Griffin2025}.
To capture the temporal dimension, researchers have investigated the scanpath prediction task, which involves predicting a sequence of fixations for a given image. Scan path prediction has historically involved generating a single deterministic output, which is insufficient to capture the inherent variability of human attention. Figure~\ref{fig:heatmap-scanpath-cont} illustrates the variation in both scanpaths (c) and eye movement trajectories (d) for a population of viewers, with each color representing a different subject. As a result, recent work has shifted towards generative modeling to better reflect the stochastic nature of eye movements. 
However, existing approaches to the generative modeling of visual attention suffer from two major limitations: First, most prior methods operate directly on scanpath data rather than on eye movement trajectories, resulting in the loss of valuable information for spatio-temporal modeling. For example, in the MIT1003 dataset~\cite{Judd2009}, the average scanpath consists of 8.4 $\pm$ 2.6 timesteps, while the raw trajectories average 723.7 $\pm$ 13.4 timesteps, indicating a substantial reduction in spatio-temporal information.
Second, many existing methods model scanpath variability either autoregressively by sampling intermediate outputs, such as saliency maps derived from fixation history~\cite{Kmmerer2022deepgazeiii, Yang2024hat, chen2021predicting}, or perform deterministic scanpath prediction~\cite{ Mondal2023gazeformer}, both of which fall short in accurately capturing the true distribution of gaze behavior.


In this paper, we hypothesize that generative modeling using the full eye movement trajectories produced by eye tracking can yield more effective representations of visual attention dynamics, in comparison to prior works that operate on scanpath data. In particular, we demonstrate that \emph{training on eye movement trajectories results in increased accuracy in scanpath prediction.} 
In addition, we argue that generative models are better suited than discriminative models for learning from this rich temporal information, particularly given the variability present in both visual tasks and data collection processes. To evaluate these hypotheses, we introduce \texttt{DiffEye}, a diffusion-based framework that is conditioned on a given visual stimulus and trained directly on raw eye movement trajectories to generate realistic eye-tracking samples. These generated trajectories can then be transformed into scanpaths or saliency maps. To enhance stimulus conditioning and improve the interaction between trajectories and the image via cross-attention, we propose a novel positional embedding strategy called \textit{Corresponding Positional Embedding (CPE)}. This method strengthens conditioning by aligning gaze information with the semantic features of the stimulus. In addition, we use high resolution image features to improve semantic precision, leading to more accurate trajectory generation. 



In summary, our key contributions are as follows:
\begin{itemize}
    \item We present the first diffusion-based training framework that directly utilizes the raw eye movement trajectories obtained from standard eye tracking datasets. Our method outperforms existing approaches in scanpath generation on unseen datasets, verifying its generalizability, and enables the synthesis of continuous eye-tracking data for natural images. 
    \item We introduce a novel positional embedding strategy, \textit{Corresponding Positional Embedding (CPE)}, which improves conditioning on visual stimuli by aligning spatial locations in both the image and trajectory space. Notably, our model achieves strong results even when trained on relatively small datasets.
    \item Unlike recent models that rely on autoregressive sampling to capture scanpath variability or produce a single deterministic prediction, our approach leverages diffusion to model the inherent variability in human gaze behavior across subjects. As a result, it can generate eye movement trajectories that can be converted into scanpaths of variable lengths, producing diverse and plausible patterns that reflect the natural variability in human attention.
\end{itemize}

\section{Related Works}
\noindent \paragraph{Eye Movement Trajectory Generation} The closest related work to this paper is DiffGaze~\cite{Jiao2024diffgaze}, which adopts a diffusion-based approach to generating eye movement trajectories in 360$^\circ$ images. While the focus on 360$^\circ$ images is innovative, the resulting models cannot be applied to the analysis of the standard RGB images used in eye tracking research. Moreover, the conditioning approach in DiffGaze relies on a single global feature vector obtained via spherical convolution. In contrast, our approach enhances conditioning through the use of CPE and patch level image features. Our ablation studies (see Table~\ref{tab:merged-ablations}) demonstrate the importance of these elements for achieving strong performance. Unfortunately, a direct comparison to DiffGaze is not possible, since the code and pretrained models are not publicly available. We hope that the release of our code and models, along with training and testing splits for eye movement trajectory data from standard datasets, will spur additional research on this understudied problem.

\noindent \paragraph{Scanpath Modeling}

Works on scanpath generation operate directly on the sequences of fixations that are extracted by postprocessing eye-tracking data~\cite{kummerer2021state}. Given an input image, these models predict the location and sequence of fixations comprising the generated scanpath. 
Previous works have explored a variety of methods and problem formulations. 
IOR-ROI~\cite{sun2019visual} uses LSTMs to predict scanpaths but outputs scanpaths of fixed length which limits its generalizability. DeepGaze III~\cite{Kmmerer2022deepgazeiii} samples from saliency maps using the fixation history, and as a result it cannot utilize the global image representation in predicting each fixation. Gazeformer~\cite{Mondal2023gazeformer} introduces ZeroGaze for unseen targets, but their deterministic method cannot produce a distribution of trajectories for a given stimulus. Some works on scanpath generation have addressed the fact that gaze behavior is dependent upon the task. Similar to us, PathGAN~\cite{assens2018pathgan} uses GANs~\cite{goodfellow2020generative} for generative modeling in the FV case, though training is unstable. At present, the only available datasets containing eye movement trajectories address the standard free viewing (FV) task. Therefore, an investigation of the performance of our DiffEye approach on non-FV tasks will depend on future data collection efforts. In contrast, Chen et al.~\cite{chen2021predicting} use reinforcement learning for target present (TP) and visual QA tasks. Other methods~\cite{rashidi2020optimal, yang2022target} support TP and the related target absent (TA) condition, but not FV. HAT~\cite{Yang2024hat} is unique in covering the three tasks FV, TP, and TA with a single model via a task query, but follows a similar sampling strategy as DeepGaze III to generate scanpaths.
In summary, no prior scanpath generation work has taken advantage of the benefits of diffusion models, and competing methods for the FV task have a variety of limitations. Our experiments (see Table~\ref{scanpath-quantitative}) compare scanpaths extracted from \texttt{DiffEye} to prior scanpath generation methods and demonstrate superior performance across a variety of metrics. We also note that our approach of directly generating eye movement trajectories gives end users the flexibility to define and extract fixations using any desired approach.

There have been a number of prior scanpath generation works that target 360$^\circ$ images. Similar to the discussion of DiffGaze above, these methods are not directly comparable to ours and are included here for the sake of completeness.
SaltiNet~\cite{assens2017saltinet} samples from saliency volumes; Pathformer3D~\cite{quan2024pathformer3d} uses transformers for autoregressive prediction. ScanGAN360~\cite{martin2022scangan360} and Scantd~\cite{wang2024scantd} apply GANs and diffusion to the 360$^\circ$ modeling task.
Finally, we note that most prior works including ours have used aggregate eye tracking data across multiple users to model attention at the population level. In contrast, a recent work from Chen et. al.~\cite{chen2024individual} develops an approach to scanpath prediction for a single user. The specialization of \texttt{DiffEye} to the 360$^\circ$ image and individual prediction tasks remain as interesting subjects for future research.

\noindent \paragraph{Saliency Modeling}
Prior works on saliency prediction have benefited from a deep connection to research in visual perception~\cite{Treisman1980, Koch1987, findlay2001visual, yarbus2013eye, zelinsky2008theory}. Itti's seminal work~\cite{Itti1998} later drew interest from the computer vision community~\cite{berg2009free, borji2012state, li2014secrets, jiang2015salicon, jetley2016end, kruthiventi2017deepfix, cornia2018predicting, Kümmerer2014deepgazeI, Linardos2021, Lou2022transalnet, Hosseini2025sum}. As saliency methods capture only spatial attention and ignore the temporal dynamics central to eye movements, they are not able to address the needs of the diverse applications that motivate this work. 
Saliency maps can be derived post hoc from the output of our \texttt{DiffEye} method via fixation extraction~\cite{salvucci2000identifying} and spatial aggregation. In this regard, \texttt{DiffEye} can be viewed as an alternative approach to saliency prediction based on the detailed modeling of gaze behavior. However, methods that are designed to directly optimize saliency prediction performance are capable of higher accuracy than an indirect approach based on fixations. Further, the saliency baselines we compared against \cite{Kümmerer2014deepgazeI, Linardos2021, Lou2022transalnet, Hosseini2025sum} were trained on combinations of various saliency datasets \cite{Judd2009, jiang2015salicon, borji2015cat2000, xu2014predicting, jiang2022does, jiang2023ueyes}, including MIT1003, while our method is only trained on the relatively smaller MIT1003 dataset.



\section{Methods}

\subsection{Preliminaries}
Given a dataset \( \mathcal{D} = \{ (I^i, \{ R_j^i \}_{j=1}^{K^i}) \}_{i=1}^{N} \), where each pair consists of a visual stimulus represented by an RGB image \( I^i \in \mathbb{R}^{H \times W \times 3} \) and a set of fixed length sequences \( R_j^i \in \mathbb{R}^{L \times 2} \) of continuous eye movement trajectories collected from \( K^i \) different subjects for the \( i ^{\text{th}}\) stimulus, our goal is to model the joint distribution of stimuli and trajectories. Each time step in \( R_j^i \) corresponds to a two-dimensional coordinate \((x, y)\), representing the spatial location of gaze at that timestep. For simplicity, we denote the set of stimuli as \( \mathcal{S} \) and the set of eye movement trajectories as \( \mathcal{R} \), and use \( R \) to refer to a single eye movement trajectory and \( I \) for a single visual stimulus.
We model the distribution over \( \mathcal{R} \) conditioned on \( \mathcal{S} \) using a diffusion model. This framework enables learning the underlying generative process of eye movement behavior in response to complex visual stimuli, capturing both temporal dynamics and spatial variability. To achieve this, \texttt{DiffEye} adopts the denoising diffusion probabilistic (DDPM) model ~\cite{Ho2020} to learn a distribution \( p_\theta(\mathcal{R} \mid \mathcal{S}) \) that approximates the true conditional distribution \( q(\mathcal{R} \mid \mathcal{S}) \) of eye-tracking data. DDPM consists of a forward process that progressively adds Gaussian noise to the trajectories over \( T_{\text{diff}} \) diffusion steps, and a reverse process that gradually denoises samples to recover realistic trajectories. The diffusion model is trained by minimizing the denoising objective \( \min_{\theta} \mathbb{E}_{t_{\text{diff}}, R^{(0)}, \epsilon} \left[ \| \epsilon - \epsilon_\theta(R^{(t_{\text{diff}})}, t_{\text{diff}}, I) \|^2 \right] \), where \( \epsilon \sim \mathcal{N}(0, \mathbf{I}) \) is the ground-truth noise and \( \epsilon_\theta(R^{(t_{\text{diff}})}, t_{\text{diff}}, I) \) is the model's prediction of noise given the noised trajectory \( R^{(t_{\text{diff}})} \), diffusion timestep \( t_{\text{diff}} \), and conditioning image \( I \).

\begin{figure*}[t]
    \centering
    \includegraphics[width=0.8\linewidth]{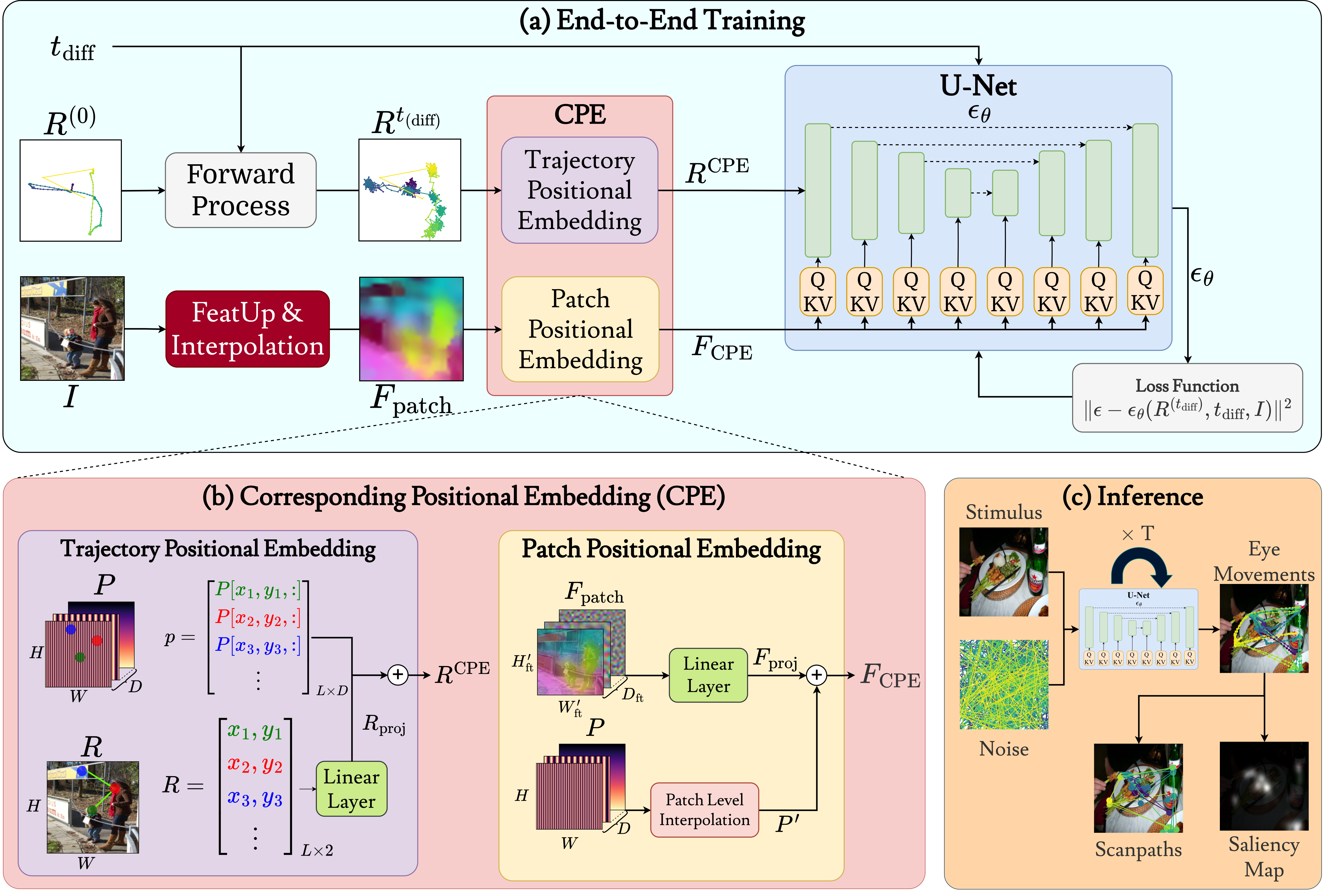}
    \caption{\textbf{\textit{An illustration of DiffEye.}} 
\textbf{(a) End-to-end Training.} Given an initial trajectory \( R^{(0)} \) and image \( I \), noise is added to produce \( R^{t_{\text{diff}}} \). FeatUp extracts patch features \( F_{\text{patch}} \), and both inputs are passed to the \textbf{(b) CPE} module, which aligns trajectory and patch positions. The resulting representations \( R^{\text{CPE}} \) and \( F_{\text{CPE}} \) are processed by a U-Net with cross-attention at each block and optimized via diffusion loss. 
\textbf{(c) Inference.} Starting from noise, the model denoises for \( T \) steps to generate an eye movement trajectory, which can be used to produce scanpaths or saliency maps.
}

    \label{fig:architecture}
\end{figure*}

\subsection{Our Approach} 


\noindent \paragraph{Base Model} We use a U-Net-based architecture \cite{Ronneberger2015} as the noise prediction model in our diffusion pipeline. The model consists of downsampling, mid, and upsampling blocks, where 1D convolution layers temporally downsample or upsample the trajectories while increasing or decreasing their number of features. Self-attention layers are applied after the convolutions to capture temporal dependencies in the eye-tracking data at multiple resolutions throughout the network. In addition to the original U-Net design, we incorporate the current diffusion timestep \( t_{\text{diff}} \) by passing it through a sinusoidal positional embedding module \cite{Vaswani2017}, followed by a fully connected layer, a SiLU activation \cite{hendrycks2016gaussian}, and another fully connected layer. The resulting timestep embedding is added to the output of the first convolution layer in each block, allowing the network to condition on \( t_{\text{diff}} \) and enabling effective diffusion training.
\noindent \paragraph{Stimuli Conditioning} \label{stimuli-conditioning}
An essential component of our pipeline is conditioning the model on a given stimulus in the form of an image. To achieve this, first, the input trajectory is passed through a 1D convolution layer, projecting it from \( R \in \mathbb{R}^{L \times 2} \) to \( R_{\text{proj}} \in \mathbb{R}^{L \times D} \), where \( L \) is the number of trajectory timesteps and \( D \) is the embedding dimension.

Initially, we attempt to condition the model using the global feature vectors produced by the DINOv2~\cite{oquab2024} Vision Transformer foundation model, chosen for its strong and well-established semantic representation capabilities. This global vector, originally of size \( \mathbb{R}^{1 \times D_{\text{global}}} \), is projected via a fully connected layer to \( \mathbb{R}^{1 \times D} \) and then concatenated with the projected trajectory tokens to form an input sequence \( E_{\text{global}} \in \mathbb{R}^{(L + 1) \times D} \). However, this approach alone leads to poor generation quality, as the global feature lacks localized spatial semantics necessary for effective conditioning.
To better incorporate localized spatial information, we utilize patch-level features from DINOv2. Let the patch features be denoted as \( F_{\text{patch}} \in \mathbb{R}^{H' \times W' \times D_{\text{feat}}} \), where \( (H', W') \) are the spatial dimensions of the patch grid. These features are projected via a linear transformation to \( F_{\text{proj}} \in \mathbb{R}^{H' \times W' \times D} \). Rather than concatenating these directly with the trajectory, we adopt a cross-attention mechanism that enables interaction between the projected image patch tokens and the trajectory tokens \( R_{\text{proj}} \in \mathbb{R}^{L \times D} \). Cross-attention is performed between the \( L \) trajectory tokens and the \( N_p = H' \cdot W' \) image patch tokens (flattened from \( F_{\text{proj}} \)). Initial attempts with a single cross-attention layer before passing the trajectory to the U-Net proved insufficient for effective conditioning. To address this, we add cross-attention layers at the end of each U-Net block, which is empirically proven to enhance the performance. Additionally, to improve spatial precision, we replace DINOv2 patch embeddings with those from FeatUp, a model-agnostic framework that restores fine-grained spatial details in deep features~\cite{fu2024featup}. FeatUp outputs a $224 \times 224 \times D_{\text{ft}}$ feature map, which we interpolate to $H'_{\text{ft}} \times W'_{\text{ft}} \times D_{\text{ft}}$, where $H'_{\text{ft}} = W'_{\text{ft}} = 32$ effectively doubling the spatial resolution relative to standard DINOv2 features.

\begin{table*}
\caption{\textbf{\textit{Training data comparison across methods.}} \texttt{DiffEye} is trained solely on the MIT1003 dataset using full eye trajectory data, while other methods are typically trained on scanpaths and rely on larger or combined datasets.}

\label{tab:training_data_comparison}
\centering
\resizebox{0.7\textwidth}{!}{%
\begin{tabular}{l>{\centering\arraybackslash}p{0.45\textwidth}ccc}
  \toprule
  \textbf{Method} & \textbf{Datasets Used} & 
  \textbf{\# Eye Trajectories} & \textbf{\# Scanpaths} & \textbf{\# Images} \\
  \midrule
  IOR-ROI & OSIE \cite{xu2014predicting} & - & 8,400 & 560 \\
  DeepGazeIII & MIT1003 \cite{Judd2009} and SALICON \cite{jiang2015salicon} & - & 615,045 & 11,003 \\
  Chen et al. & AiR \cite{air} & - & 39,080 & 1,454 \\
  GazeFormer & COCO-Search18 \cite{chen2021coco} & - & 62,020 & 6,202 \\
  HAT & \begin{tabular}[c]{@{}c@{}}COCO-Search18 \cite{chen2021coco}, COCO-FreeView \cite{chen2022characterizing},\\ MIT1003 \cite{Judd2009}, OSIE \cite{xu2014predicting}\end{tabular} & - & 87,565 & 7,905 \\
  \texttt{DiffEye} & MIT1003 \cite{Judd2009} & 8,934 & - & 1,003 \\
  \bottomrule
\end{tabular}
}
\end{table*}

\noindent \paragraph{Corresponding Positional Embedding (CPE)}  
To further improve spatial alignment, we introduce a novel positional embedding method, \textit{Corresponding Positional Embedding (CPE)}. Let the $t_\text{diff}$-step noisified input trajectory \( R^{t_{(\text{diff})}} \) be denoted as \( R \in \mathbb{R}^{L \times 2} \), where each row represents a 2D coordinate \( (x_i, y_i) \). We first project the trajectory into the hidden dimension, resulting in the embedded trajectory \( R_{\text{proj}} \in \mathbb{R}^{L \times D} \). Next, we construct a 2D positional embedding grid \( P \in \mathbb{R}^{H \times W \times D} \), based on the original image resolution \( (H, W) \), using sinusoidal encodings~\cite{Vaswani2017}. For each timestep $i\in \{1,L\}$, we retrieve the corresponding positional embedding vector from the grid as \( p_i = P[y_i, x_i, :] \in \mathbb{R}^{D} \), where \( R_i = (x_i, y_i) \) is the $i$-th coordinate in the original trajectory \( R \). The final CPE-augmented trajectory token is then computed as:
\begin{align}
R_i^{\text{CPE}} = R_{\text{proj}}[i] + p_i.
\end{align}
The same positional grid \( P \) is also applied to the image features. Let the projected image features be \( F_{\text{proj}} \in \mathbb{R}^{H' \times W' \times D} \). We interpolate the positional grid \( P \) to the patch resolution, yielding \( P' \in \mathbb{R}^{H' \times W' \times D} \). We then add this to the image features:
\begin{align}
F_{\text{CPE}} = F_{\text{proj}} + P'.
\end{align}
By sharing and aligning positional information across both trajectories and image patches, CPE enables effective spatial correspondence and interaction between gaze behavior and visual content. Figure \ref{fig:architecture} summarizes our approach, including end-to-end training (a), the CPE module (b), and inference (c).
\noindent \paragraph{Data Preprocessing} \label{sec:data_preprocessing}
We use the MIT1003 dataset~\cite{Judd2009} \footnote{\url{https://people.csail.mit.edu/tjudd/WherePeopleLook/index.html}} to train and evaluate our model. To the best of our knowledge, it is the only publicly available dataset that provides raw eye-tracking data for natural images obtained during a free-viewing task. The dataset contains recordings from 15 subjects who free-viewed 1,003 images for 3 seconds each, resulting in a total of 15{,}045 eye-tracking sequences. Data was collected using the ETL 400 ISCAN system at a sampling rate of 240~Hz. During preprocessing, we removed all blinks and NaN values, and retained only the sequences with at least 720 timesteps. After filtering, 8{,}934 trajectories remained. All remaining trajectories were truncated or downsampled to a uniform length of 720 samples, as our model is designed to generate fixed-length eye movement trajectories and the U-Net architecture requires several downsampling steps which is possible with a sequence length of 720. Additionally, all stimuli were resized to $224 \times 224$ pixels. The dataset is split into training (90\%) and testing (10\%) sets based on stimuli, ensuring that images used for evaluation were not seen during training. 

\section{Experimentation} \label{sec:experimentation}

\begin{figure*}
    \centering
    \includegraphics[width=0.75\linewidth]{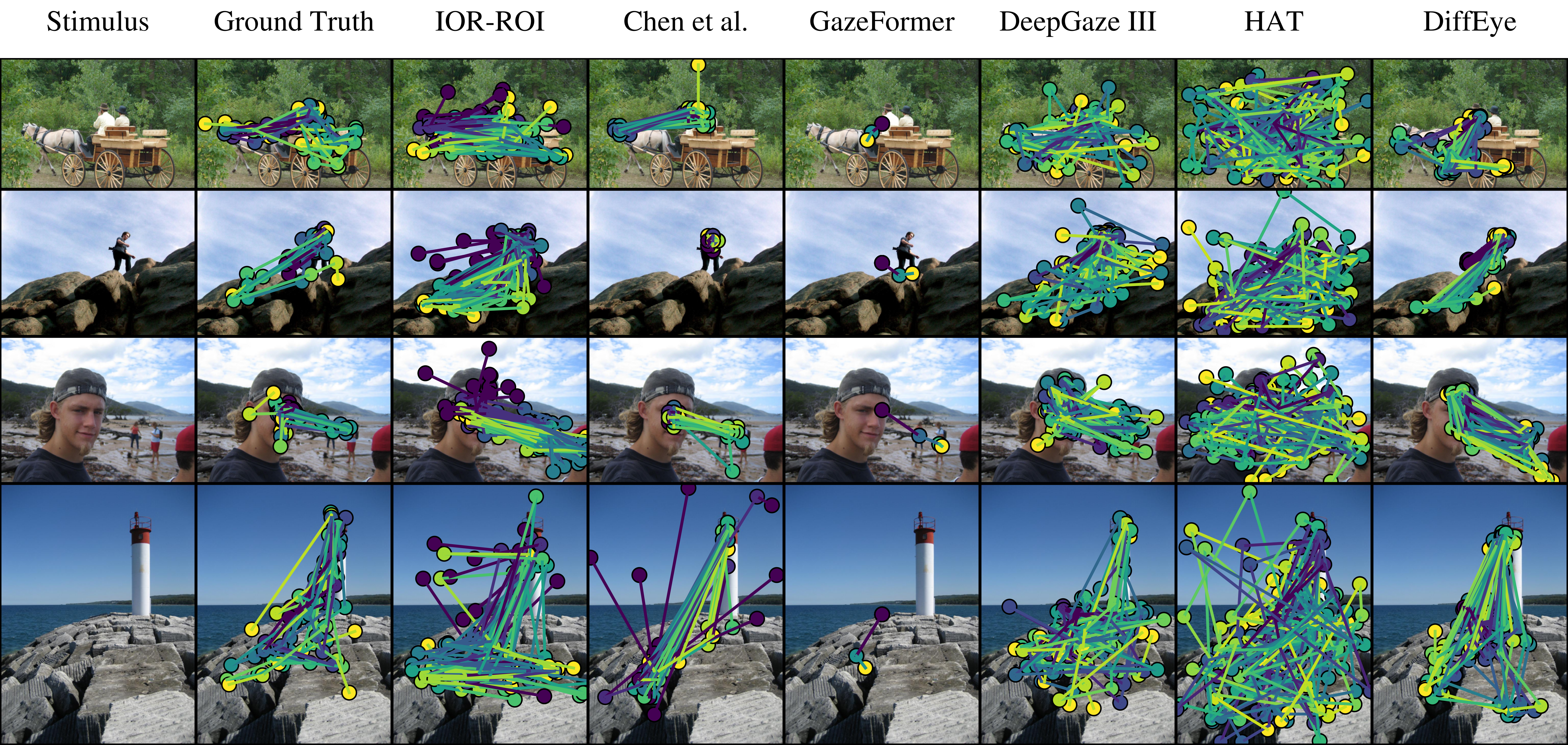}
\caption{\textbf{\textit{Qualitative comparison of scanpath generation.}} 
Scanpaths generated by \texttt{DiffEye} and baseline models are shown alongside ground truth annotations across four different scenes. 
Each row represents a unique stimulus, and each column shows the corresponding scanpaths generated from a specific method.}

    \label{fig:scanpath-qualitative}
\end{figure*}

\noindent \paragraph{Implementation Details} \label{training_and_sampling}
We train our model using the Adam optimizer~\cite{kingma2014adam} for 3000 epochs with a fixed learning rate of \(1 \times 10^{-4}\). During training, we utilize the DDPM scheduler with a linear noise schedule ranging from \(1 \times 10^{-4}\) to \(2 \times 10^{-2}\). For sampling, we adopt Denoising Diffusion Implicit Models (DDIM)~\cite{song2021ddim}, which enable high-quality sample generation in significantly fewer steps without compromising performance. We set the number of diffusion steps to \( T_{\text{diff}} = 1000 \) during training and reduce it to 50 during sampling for improved efficiency. Throughout both training and inference, we apply classifier-free guidance (CFG) to effectively condition the model on the visual stimulus \cite{ho2021classifier}. This encourages the model to learn both conditional and unconditional denoising behavior, supporting effective guidance at inference time. See Supplementary for the complete details. All experiments were conducted using an NVIDIA RTX A6000 GPU. 


\begin{table*}[t]
\caption{\textbf{\textit{Ablation study for scanpath and eye movement trajectory generation.}} 
We evaluate the contribution of key components in \texttt{DiffEye} by removing modules such as FeatUp, CPE (Corresponding Positional Embedding), U-Net cross-attention, and patch-level features. 
Results are reported for both scanpath and continuous trajectory generation using four trajectory-based evaluation metrics.}

\label{tab:merged-ablations}
\centering
\resizebox{0.7\textwidth}{!}{%
\begin{tabular}{llcccccccc}
  \toprule
  \textbf{Task} & \textbf{Configuration} 
  & \multicolumn{2}{c}{\begin{tabular}{@{}c@{}}Levenshtein\\Distance ($\times 10^2$) $\downarrow$\end{tabular}} 
  & \multicolumn{2}{c}{\begin{tabular}{@{}c@{}}Discrete Fréchet\\Distance ($\times 10^2$) $\downarrow$\end{tabular}} 
  & \multicolumn{2}{c}{\begin{tabular}{@{}c@{}}Dynamic Time\\Warping ($\times 10^4$) $\downarrow$\end{tabular}} 
  & \multicolumn{2}{c}{\begin{tabular}{@{}c@{}}Time Delay\\Embedding $\downarrow$\end{tabular}} \\
  
  \cmidrule(lr){3-4}
  \cmidrule(lr){5-6}
  \cmidrule(lr){7-8}
  \cmidrule(lr){9-10}
  
  & & \textit{Mean} & \textit{Best} 
    & \textit{Mean} & \textit{Best} 
    & \textit{Mean} & \textit{Best} 
    & \textit{Mean} & \textit{Best} \\
  
  \midrule
  \multirow{5}{*}{\begin{tabular}[c]{@{}l@{}}\textbf{Scanpath}\\\textbf{Generation}\end{tabular}} 
  & \texttt{DiffEye} & \cellcolor{gray!10}\textbf{0.130} & \cellcolor{gray!30}\textbf{0.097} & \cellcolor{gray!10}\textbf{3.529} & \cellcolor{gray!30}\textbf{2.449} & \cellcolor{gray!10}\textbf{0.157} & \cellcolor{gray!30}\textbf{0.107} & \cellcolor{gray!10}\textbf{88.661} & \cellcolor{gray!30}\textbf{53.486} \\
  & \textbf{w/o} FeatUp & \cellcolor{gray!10}0.133 & \cellcolor{gray!30}0.100 & \cellcolor{gray!10}3.546 & \cellcolor{gray!30}2.423 & \cellcolor{gray!10}0.163 & \cellcolor{gray!30}0.110 & \cellcolor{gray!10}91.007 & \cellcolor{gray!30}60.103 \\
  & \textbf{w/o} CPE & \cellcolor{gray!10}0.141 & \cellcolor{gray!30}0.107 & \cellcolor{gray!10}3.545 & \cellcolor{gray!30}2.604 & \cellcolor{gray!10}0.180 & \cellcolor{gray!30}0.128 & \cellcolor{gray!10}100.792 & \cellcolor{gray!30}69.827 \\
  & \textbf{w/o} U-Net Cross-Attention & \cellcolor{gray!10}0.143 & \cellcolor{gray!30}0.107 & \cellcolor{gray!10}3.701 & \cellcolor{gray!30}2.557 & \cellcolor{gray!10}0.189 & \cellcolor{gray!30}0.130 & \cellcolor{gray!10}107.962 & \cellcolor{gray!30}68.353 \\
  & \textbf{w/o} Patch Level Features & \cellcolor{gray!10}0.153 & \cellcolor{gray!30}0.116 & \cellcolor{gray!10}3.761 & \cellcolor{gray!30}2.692 & \cellcolor{gray!10}0.209 & \cellcolor{gray!30}0.147 & \cellcolor{gray!10}116.226 & \cellcolor{gray!30}77.997 \\
  
  \midrule
  \multirow{5}{*}{\begin{tabular}[c]{@{}l@{}}\textbf{Eye Movement}\\\textbf{Trajectory Generation}\end{tabular}} 
  & \texttt{DiffEye} & \cellcolor{gray!10}\textbf{10.083} & \cellcolor{gray!30}\textbf{8.289} & \cellcolor{gray!10}\textbf{3.601} & \cellcolor{gray!30}\textbf{2.460} & \cellcolor{gray!10}\textbf{11.834} & \cellcolor{gray!30}\textbf{8.212} & \cellcolor{gray!10}\textbf{35.228} & \cellcolor{gray!30}\textbf{20.968} \\
  & \textbf{w/o} FeatUp & \cellcolor{gray!10}10.265 & \cellcolor{gray!30}8.736 & \cellcolor{gray!10}3.844 & \cellcolor{gray!30}2.623 & \cellcolor{gray!10}12.513 & \cellcolor{gray!30}8.645 & \cellcolor{gray!10}41.224 & \cellcolor{gray!30}26.453 \\
  & \textbf{w/o} CPE & \cellcolor{gray!10}10.773 & \cellcolor{gray!30}9.200 & \cellcolor{gray!10}3.621 & \cellcolor{gray!30}2.599 & \cellcolor{gray!10}13.430 & \cellcolor{gray!30}10.068 & \cellcolor{gray!10}44.403 & \cellcolor{gray!30}28.904 \\
  & \textbf{w/o} U-Net Cross-Attention & \cellcolor{gray!10}10.971 & \cellcolor{gray!30}9.394 & \cellcolor{gray!10}3.828 & \cellcolor{gray!30}2.587 & \cellcolor{gray!10}14.716 & \cellcolor{gray!30}10.992 & \cellcolor{gray!10}56.739 & \cellcolor{gray!30}38.264 \\
  & \textbf{w/o} Patch-Level Features & \cellcolor{gray!10}11.791 & \cellcolor{gray!30}9.947 & \cellcolor{gray!10}4.088 & \cellcolor{gray!30}2.761 & \cellcolor{gray!10}18.007 & \cellcolor{gray!30}13.312 & \cellcolor{gray!10}77.042 & \cellcolor{gray!30}47.354 \\
  
  \bottomrule
\end{tabular}
}
\end{table*}

\begin{figure*}[t]
    \centering
    \includegraphics[width=0.775\linewidth]{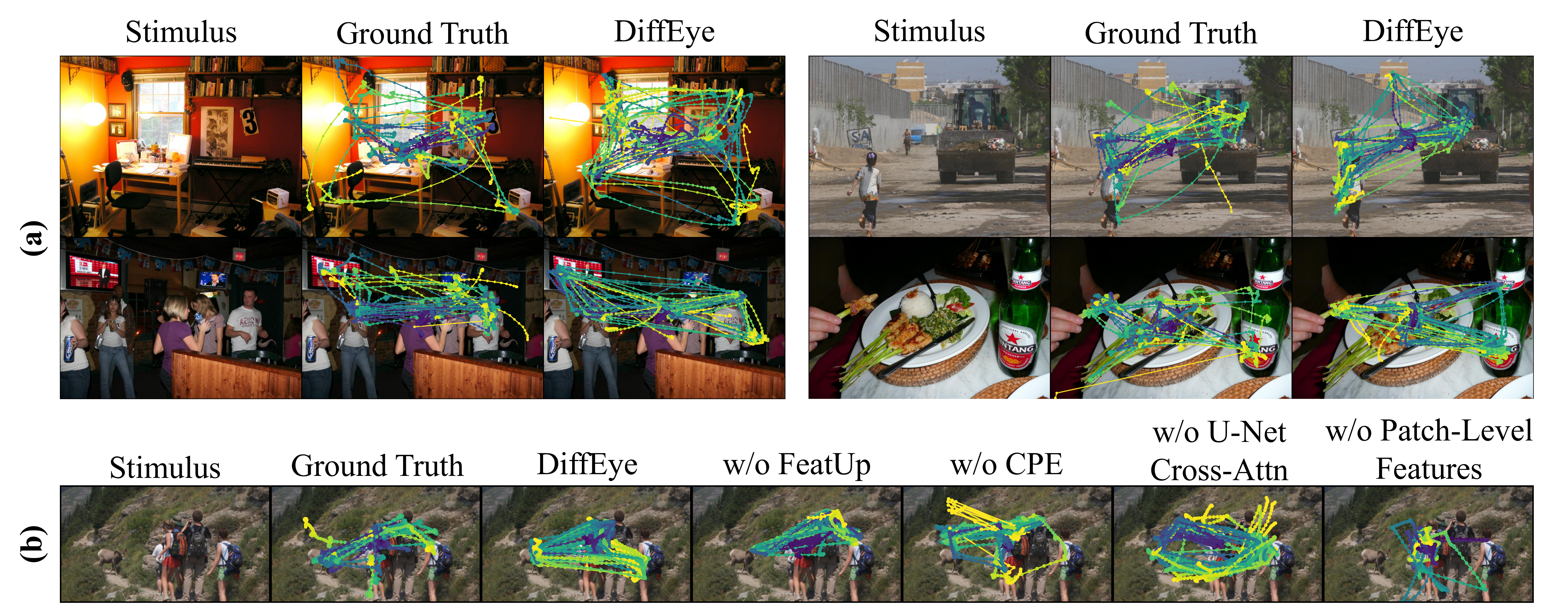}
\caption{\textbf{\textit{Qualitative analysis and ablation study of continuous eye movement trajectory generation.}} 
(a) Multiple eye movement trajectories generated by \texttt{DiffEye}) alongside ground truth annotations across four different scenes. 
(b) Ablation study showing the impact of removing individual architectural components (FeatUp, CPE, cross-attention, and patch-level features) on continuous trajectory generation.}
    \label{cont_ablations_qual}
\end{figure*}

\noindent \paragraph{Baselines and Datasets} \label{scanpath-generation-method} We compare our method against several baseline models trained for the scanpath prediction/generations tasks. The baselines include the Human Attention Transformer (HAT)~\cite{Yang2024hat} and GazeFormer~\cite{Mondal2023gazeformer}, both of which are transformer-based architectures, DeepGaze III~\cite{Kmmerer2022deepgazeiii}, a convolutional network, Chen et al. \cite{chen2021predicting}, which is a reinforcement learning model that mainly focuses on visual question answering, and IOR-ROI \cite{sun2019visual}. Note that GazeFormer can not generate a distribution but can only predict scanpaths. DeepGaze III and IOR-ROI requires the number of fixations per scanpath as an input, which we set as 10 for all experiments. Each baseline is used to generate scanpaths for all test stimuli. Scanpath baselines were trained on scanpaths coming from various combinations of datasets as shown in Table \ref{tab:training_data_comparison}. In contrast, our method is trained solely on a subset of the raw eye movement trajectories from the MIT1003 dataset which is relatively very small dataset with fewer pairs of stimuli and eye movement trajectories, yet achieves competitive performance, demonstrating robustness under limited data. We evaluate our results on the test sets of the MIT1003 and OSIE \cite{xu2014predicting} \footnote{\url{https://github.com/chenxy99/Scanpaths/tree/main/OSIE}} datasets. It is important to note that the MIT1003 dataset does not provide official train-test splits; therefore, we created a random split of 100 images with eye-tracking data from 15 subjects each as described in Section \ref{sec:data_preprocessing}. As a result, methods such as HAT and DeepGazeIII, whose training data included MIT1003, have already been exposed to the evaluation data. The OSIE test set contains 70 images with eye-tracking data from 15 subjects.

\begin{table*}
\caption{\textbf{\textit{Quantitative comparison of scanpath generation on the MIT1003 and OSIE datasets.}} 
We report the performance of each method using four commonly used trajectory-based metrics: Levenshtein Distance, Discrete Fréchet Distance, Dynamic Time Warping (DTW), and Time Delay Embedding (TDE). The label (seen) indicates that the model's training set includes the test set images.
\textbf{Bold} values indicate the best scores, while \underline{underlined} values denote the second-best scores.}
\label{scanpath-quantitative}
\centering
\resizebox{0.7\textwidth}{!}{%
\begin{tabular}{llc c c c c c c c c}
  \toprule
  Test Dataset & Method 
  & \multicolumn{2}{c}{\begin{tabular}{@{}c@{}}Levenshtein\\Distance $\downarrow$\end{tabular}} 
  & \multicolumn{2}{c}{\begin{tabular}{@{}c@{}}Discrete Fréchet\\Distance $\downarrow$ ($\times 10^2$)\end{tabular}} 
  & \multicolumn{2}{c}{\begin{tabular}{@{}c@{}}Dynamic Time\\Warping $\downarrow$ ($\times 10^3$)\end{tabular}} 
  & \multicolumn{2}{c}{\begin{tabular}{@{}c@{}}Time Delay\\Embedding $\downarrow$\end{tabular}} \\
  
  \cmidrule(lr){3-4}
  \cmidrule(lr){5-6}
  \cmidrule(lr){7-8}
  \cmidrule(lr){9-10}
  
  & & \textit{Mean} & \textit{Best} 
  & \textit{Mean} & \textit{Best} 
  & \textit{Mean} & \textit{Best} 
  & \textit{Mean} & \textit{Best} \\
  
  \midrule
  \multirow{6}{*}{MIT1003}
  & IOR-ROI & \cellcolor{gray!10}\underline{13.574} & \cellcolor{gray!30}\underline{11.092} & \cellcolor{gray!10}3.777 & \cellcolor{gray!30}2.460 & \cellcolor{gray!10}1.834 & \cellcolor{gray!30}1.317 & \cellcolor{gray!10}108.284 & \cellcolor{gray!30}80.944 \\
  & DeepGaze III (seen) & \cellcolor{gray!10}14.415 & \cellcolor{gray!30}11.856 & \cellcolor{gray!10}3.553 & \cellcolor{gray!30}\textbf{2.160} & \cellcolor{gray!10}\underline{1.757} & \cellcolor{gray!30}\underline{1.141} & \cellcolor{gray!10}96.456 & \cellcolor{gray!30}\underline{65.408} \\
  & Chen et al. & \cellcolor{gray!10}14.874 & \cellcolor{gray!30}12.943 & \cellcolor{gray!10}\underline{3.704} & \cellcolor{gray!30}2.602 & \cellcolor{gray!10}1.851 & \cellcolor{gray!30}1.409 & \cellcolor{gray!10}\underline{92.100} & \cellcolor{gray!30}74.212 \\
  & GazeFormer & \cellcolor{gray!10}- & \cellcolor{gray!30}12.614 & \cellcolor{gray!10}- & \cellcolor{gray!30}3.553 & \cellcolor{gray!10}- & \cellcolor{gray!30}1.545 & \cellcolor{gray!10}- & \cellcolor{gray!30}93.751 \\
  & HAT (seen) & \cellcolor{gray!10}18.440 & \cellcolor{gray!30}14.645 & \cellcolor{gray!10}4.293 & \cellcolor{gray!30}2.940 & \cellcolor{gray!10}2.680 & \cellcolor{gray!30}1.862 & \cellcolor{gray!10}131.516 & \cellcolor{gray!30}97.232 \\
  & \texttt{DiffEye} & \cellcolor{gray!10}\textbf{13.009} & \cellcolor{gray!30}\textbf{9.709} & \cellcolor{gray!10}\textbf{3.529} & \cellcolor{gray!30}\underline{2.449} & \cellcolor{gray!10}\textbf{1.573} & \cellcolor{gray!30}\textbf{1.067} & \cellcolor{gray!10}\textbf{88.661} & \cellcolor{gray!30}\textbf{53.486} \\
  
  \midrule
  \multirow{6}{*}{OSIE} 
  & IOR-ROI & \cellcolor{gray!10}\underline{14.836} & \cellcolor{gray!30}\underline{12.152} & \cellcolor{gray!10}3.357 & \cellcolor{gray!30}\underline{2.228} & \cellcolor{gray!10}\underline{1.699} & \cellcolor{gray!30}1.167 & \cellcolor{gray!10}92.960 & \cellcolor{gray!30}70.624 \\
  & DeepGaze III & \cellcolor{gray!10}15.507 & \cellcolor{gray!30}12.532 & \cellcolor{gray!10}\underline{3.206} & \cellcolor{gray!30}\textbf{2.077} & \cellcolor{gray!10}1.765 & \cellcolor{gray!30}\underline{1.166} & \cellcolor{gray!10}84.337 & \cellcolor{gray!30}\underline{57.786} \\
  & Chen et al. & \cellcolor{gray!10}17.024 & \cellcolor{gray!30}14.910 & \cellcolor{gray!10}3.275 & \cellcolor{gray!30}2.290 & \cellcolor{gray!10}1.772 & \cellcolor{gray!30}1.276 & \cellcolor{gray!10}\textbf{78.286} & \cellcolor{gray!30}61.509 \\
  & GazeFormer & \cellcolor{gray!10}- & \cellcolor{gray!30}15.320 & \cellcolor{gray!10}- & \cellcolor{gray!30}3.257 & \cellcolor{gray!10}- & \cellcolor{gray!30}1.687 & \cellcolor{gray!10}- & \cellcolor{gray!30}81.878 \\
  & HAT & \cellcolor{gray!10}19.419 & \cellcolor{gray!30}15.607 & \cellcolor{gray!10}3.712 & \cellcolor{gray!30}2.598 & \cellcolor{gray!10}2.501 & \cellcolor{gray!30}1.757 & \cellcolor{gray!10}111.413 & \cellcolor{gray!30}83.140 \\
  & \texttt{DiffEye} & \cellcolor{gray!10}\textbf{14.771} & \cellcolor{gray!30}\textbf{12.077} & \cellcolor{gray!10}\textbf{3.068} & \cellcolor{gray!30}2.238 & \cellcolor{gray!10}\textbf{1.552} & \cellcolor{gray!30}\textbf{1.089} & \cellcolor{gray!10}\underline{81.925} & \cellcolor{gray!30}\textbf{54.347} \\
  
  \bottomrule
\end{tabular}
}
\end{table*}

\noindent \paragraph{Evaluation Metrics} For scanpath generation, we follow the evaluation protocol proposed in~\cite{wang2024scantd, Jiao2024diffgaze}. For each image in the test set, we generate 15 eye-tracking trajectories per model, extract fixation points using the conversion method described in~\cite{Judd2009}, and construct the corresponding scanpaths. For models supporting multiple visual tasks, we ensured that scanpaths were generated specifically for the Free-Viewing task.
We evaluate the results using four standard metrics commonly used in the eye-tracking literature: \textit{Levenshtein Distance}, which measures the sequence similarity based on insertion, deletion, and substitution operations; \textit{Discrete Fréchet Distance (DFD)}, which captures the similarity between curves while considering their sequential nature; \textit{Dynamic Time Warping (DTW)}, which aligns sequences that may vary in speed or length; and \textit{Time Delay Embedding (TDE)}, which assesses the temporal structure of trajectories~\cite{Fahimi2020}. Each test image is associated with 15 ground truth scanpaths. We generate 15 scanpaths per model; for deterministic models such as Gazeformer, only a single prediction is evaluated.
For each metric, we report two scores: \textit{best} and \textit{mean}. The \textit{best} score is obtained by computing the evaluation metric between each ground truth scanpath and all generated scanpaths, selecting the most similar (or least distant) one, and then averaging over all scanpaths and images. The \textit{mean} score is obtained by averaging the metric across all pairwise comparisons between ground truth and generated scanpaths for each image, followed by averaging across all test images (see Algorithm~\ref{alg:evaluation} in Supplementary.).

\noindent \paragraph{Scanpath Generation Results}
As shown in Table~\ref{scanpath-quantitative}, the \textit{mean} scores highlight our model’s superior ability to generate trajectories that align closely with the overall distribution of human fixations. This is demonstrated by the lowest Levenshtein and TDE values, which capture spatial and temporal deviations across full scanpaths. Furthermore, the DTW and DFD metric, sensitive to the shape and curvature of the trajectories, indicates that our model better preserves the sequential flow of eye movements.
Figure~\ref{fig:scanpath-qualitative} visually reinforces these findings. In the second row (the image of the climbing human), the model by Chen et al. overly concentrates predictions around the low-resolution human head, while \texttt{DiffEye} distributes gaze more naturally across salient regions, including the background, showing stronger alignment with the ground truth. In contrast, DeepGaze III and HAT produce diffuse trajectories, suggesting a failure to learn coherent attention patterns. IOR-ROI performs slightly better, but still deviates from the true distribution. Gazeformer, meanwhile, does not generate a distribution, but performs prediction, resulting in higher Levenshtein and TDE mean scores, indicating weaker alignment with human gaze distributions. Note that our model is trained on the MIT1003 dataset and evaluated on both the MIT1003 test set and the entirely unseen OSIE dataset, demonstrating strong generalization. Additional qualitative results are included in Supplementary.

\begin{figure*}
    \centering
    \includegraphics[width=.825\linewidth]{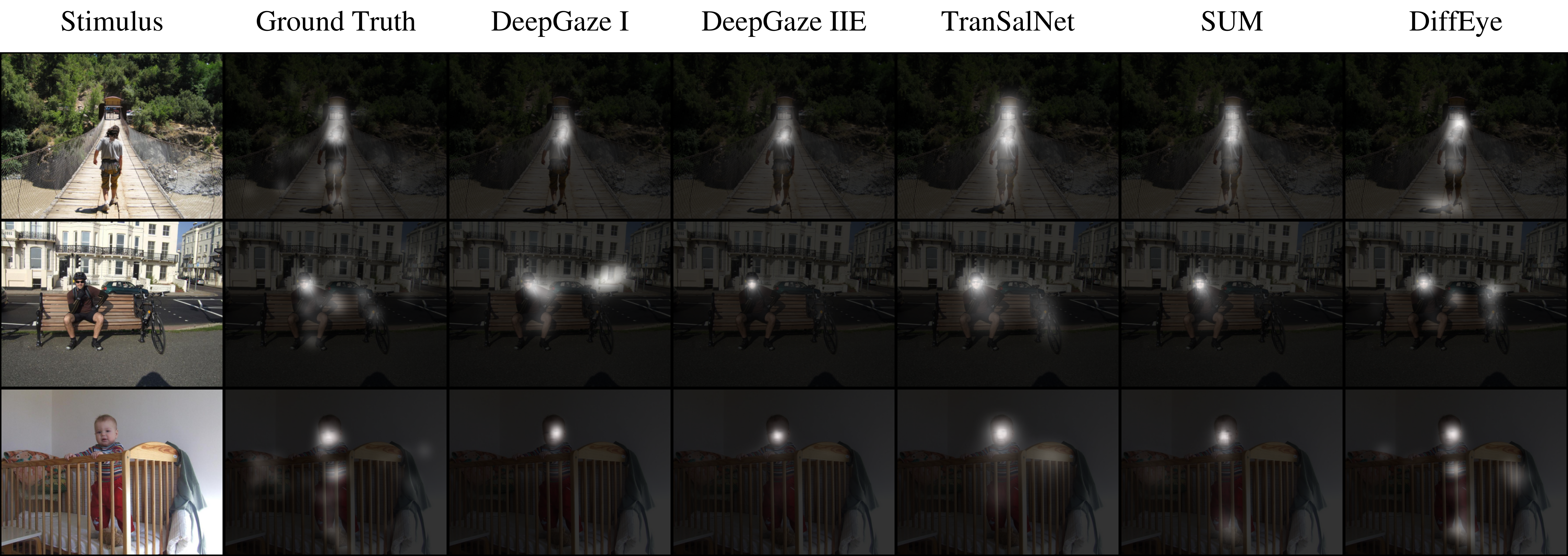}
\caption{\textbf{\textit{Qualitative comparison of saliency map predictions.}} 
Saliency maps generated by \texttt{DiffEye} and baseline models are shown alongside ground truth maps for four different scenes. 
Each row corresponds to a different stimulus, with columns displaying the stimulus, ground truth, and predictions.}

    \label{fig:saliency-qualitative}
\end{figure*}

\noindent \paragraph{Ablation Study}

We conduct an ablation study to assess the importance of key architectural components in \texttt{DiffEye}. Specifically, we compare the full model, comprising FeatUp high-resolution patch features, the proposed CPE, and cross-attention at all U-Net layers, against variants where one of these components was removed.
As shown in Table~\ref{tab:merged-ablations}, ablating any of these components degrades performance across both scanpath and continuous eye movement trajectory tasks. Removing FeatUp and replacing it with DINOv2 features leads to a consistent drop in performance. This confirms the benefit of FeatUp, which provides higher resolution and spatially structured visual features essential for precise modeling. The CPE module, a novel contribution of our work, also significantly boosts performance. By aligning eye trajectories with the spatial layout of the visual stimulus, CPE enhances the model’s ability to localize attention signals effectively, especially evident in the Levenshtein and Fréchet distances, which are sensitive to spatial alignment. We further evaluate the effect of distributing cross-attention across all U-Net blocks versus using a single cross-attention layer at the beginning. Disabling full cross-attention notably worsens performance, suggesting that deep conditioning through the network is crucial to preserve and propagate stimulus-related signals. Finally, we test using only a global token instead of patch-level features which is concatenated as an additional token to the trajectory tokens before passing it through the U-Net. This setup parallels DiffGaze~\cite{Jiao2024diffgaze}, a model designed for 360° images, which also uses a conditioning mechanism with global token. Our results show that such global conditioning is insufficient for natural images, where fine-grained patch-level information is needed to localize saliency and highlight the value of our patch-based strategy. Please refer to Figure~\ref{cont_ablations_qual} for qualitative comparisons that further illustrate the contributions of each component. See Supplementary for additional qualitative results for the eye movement trajectory generation.

\noindent \paragraph{Saliency Prediction}
We compare our method against the SUM model \cite{Hosseini2025sum}, TranSalNet \cite{Lou2022transalnet}, DeepGaze I \cite{Kümmerer2014deepgazeI}, and DeepGaze IIE \cite{Linardos2021} which are all trained specifically for the saliency prediction task. Each baseline model is used to generate saliency maps for the MIT1003 test split used to evaluate scanpaths. We convert the continuous eye movement trajectories generated from \texttt{DiffEye} to saliency maps as described in the Supplementary materials. Figure~\ref{fig:saliency-qualitative} presents qualitative results for the saliency prediction task comparing \texttt{DiffEye} with baseline models. Despite not being explicitly trained for saliency map generation, our model produces visually competitive outputs. For instance, in the second row, while DeepGaze IIE and SUM primarily focus on the subject's face, \texttt{DiffEye} captures a broader set of salient regions, including the face, the bicycle, and the left hand, closely resembling the ground truth distribution. These results demonstrate our model’s capacity to generalize beyond its original training objective. Additional evaluations and qualitative results are provided in Supplementary.

\section{Discussion \& Limitations} \label{sec:discussion_limitations}
We present \texttt{DiffEye}, a novel diffusion-based model for generating eye movement trajectories on natural images. Unlike prior methods that rely on deterministic architectures, autoregressive sampling, and training with compressed representations like saliency maps or discrete scanpaths, \texttt{DiffEye} captures the variability of human visual behavior using state-of-the-art generative modeling and trains directly on raw eye-tracking data. To support this, we introduce CPE, a mechanism that aligns spatial gaze patterns with semantic content via cross-attention between image patches and trajectory timesteps.
By modeling scanpaths as a generative process, \texttt{DiffEye} produces realistic and diverse gaze behaviors and achieves state-of-the-art results. Beyond technical performance, it has potential applications in developmental science by simulating population-specific gaze patterns. This may support the diagnosis of conditions such as autism by generating stimuli that maximize group-level distinctions in eye-tracking studies. Although \texttt{DiffEye} achieves strong results using only the MIT1003 dataset—the only known dataset offering raw eye trajectories alongside scanpaths for natural images—the dataset size remains a key limitation and scaling its performance will require access to additional data. Future work will explore transfer learning from datasets containing saliency maps and scanpaths, and aim to include data from both neurotypical and developmentally diverse populations. We also advocate for the release of raw eye-tracking data. At present, \texttt{DiffEye} generates only fixed-length outputs at 240 Hz; broader data availability would enable support for variable-length sequences and diverse sampling rates. Ultimately, \texttt{DiffEye} could also be used to synthesize training data, helping to address data scarcity in this field.

\begin{ack}
Portions of this research were supported in part by the Health Care Engineering Systems Center in the Grainger College of Engineering at UIUC, and the National Institutes of Health (NIH) under award P41EB028242.
\end{ack}



{
\small
\bibliographystyle{unsrt} 
\bibliography{references}
}

\newpage
\appendix

\section{Appendix / supplemental material}

\subsection{Classifier Free Guidance}
Given a noised eye movement trajectory \( R^{(t_{\text{diff}})} \) at diffusion timestep \( t_{\text{diff}} \), we perform two forward passes through the noise prediction network \( \epsilon_\theta \): one conditioned on the visual stimulus \( I \), yielding \( \epsilon_\theta(R^{(t_{\text{diff}})}, t_{\text{diff}}, I) \), and another using an unconditional input \( I_{\text{uc}} \), defined as a zero matrix, yielding \( \epsilon_\theta(R^{(t_{\text{diff}})}, t_{\text{diff}}, I_{\text{uc}}) \). The final guided noise prediction \( \hat{\epsilon}_\theta \) is computed as:
\begin{align}
\hat{\epsilon}_\theta = (1 - c) \cdot \epsilon_\theta(R^{(t_{\text{diff}})}, t_{\text{diff}}, I_{\text{uc}}) + c \cdot \epsilon_\theta(R^{(t_{\text{diff}})}, t_{\text{diff}}, I),
\end{align}
where \( c \) is the classifier-free guidance scale, which we set to 4 during inference. To enable this, we simulate the unconditional setting during training by randomly replacing the conditioning input \( I \) with a zero matrix in 10\% of the training samples. This encourages the model to learn both conditional and unconditional denoising behavior, supporting effective guidance at inference time.

\subsection{Evaluation Algorithm} \label{a:eval_algo}
Below is the algorithm we used to compute the \textit{mean} and \textit{best} scores for each of the scanpath and continuous trajectory metrics.

\begin{algorithm}[H]
\caption{Evaluation of Scanpath and Continuous Trajectory Generation Metrics}
\label{alg:evaluation}
\small  
\begin{algorithmic}[1]
\Require Test set of images $\mathcal{I}$; ground truth scanpaths $\mathcal{G}_i = \{g_1, \ldots, g_N\}$; generated scanpaths $\mathcal{S}_i = \{s_1, \ldots, s_M\}$; evaluation metric $d(\cdot,\cdot)$
\Ensure Overall \textit{best} and \textit{mean} scores across test images
\State Initialize \texttt{best\_scores} $\gets$ [ ] and \texttt{mean\_scores} $\gets$ [ ]
\For{each image $i \in \mathcal{I}$}
    \State Initialize \texttt{image\_best} $\gets$ [ ] and \texttt{image\_mean} $\gets$ [ ]
    \For{each $g \in \mathcal{G}_i$}
        \State Compute distances $\{d(g, s) \mid s \in \mathcal{S}_i\}$
        \State $\texttt{best\_g} \gets \min_{s \in \mathcal{S}_i} d(g, s)$
        \State $\texttt{mean\_g} \gets \frac{1}{M} \sum_{s \in \mathcal{S}_i} d(g, s)$
        \State Append \texttt{best\_g} to \texttt{image\_best}
        \State Append \texttt{mean\_g} to \texttt{image\_mean}
    \EndFor
    \State Append $\frac{1}{N} \sum \texttt{image\_best}$ to \texttt{best\_scores}
    \State Append $\frac{1}{N} \sum \texttt{image\_mean}$ to \texttt{mean\_scores}
\EndFor
\State \Return Overall Best $= \frac{1}{|\mathcal{I}|} \sum \texttt{best\_scores}$, Overall Mean $= \frac{1}{|\mathcal{I}|} \sum \texttt{mean\_scores}$
\end{algorithmic}
\end{algorithm}

\subsection{Additional Scanpath Distributions} \label{a:sp}
\begin{figure}[H]
    \centering
    \includegraphics[width=.9\linewidth]{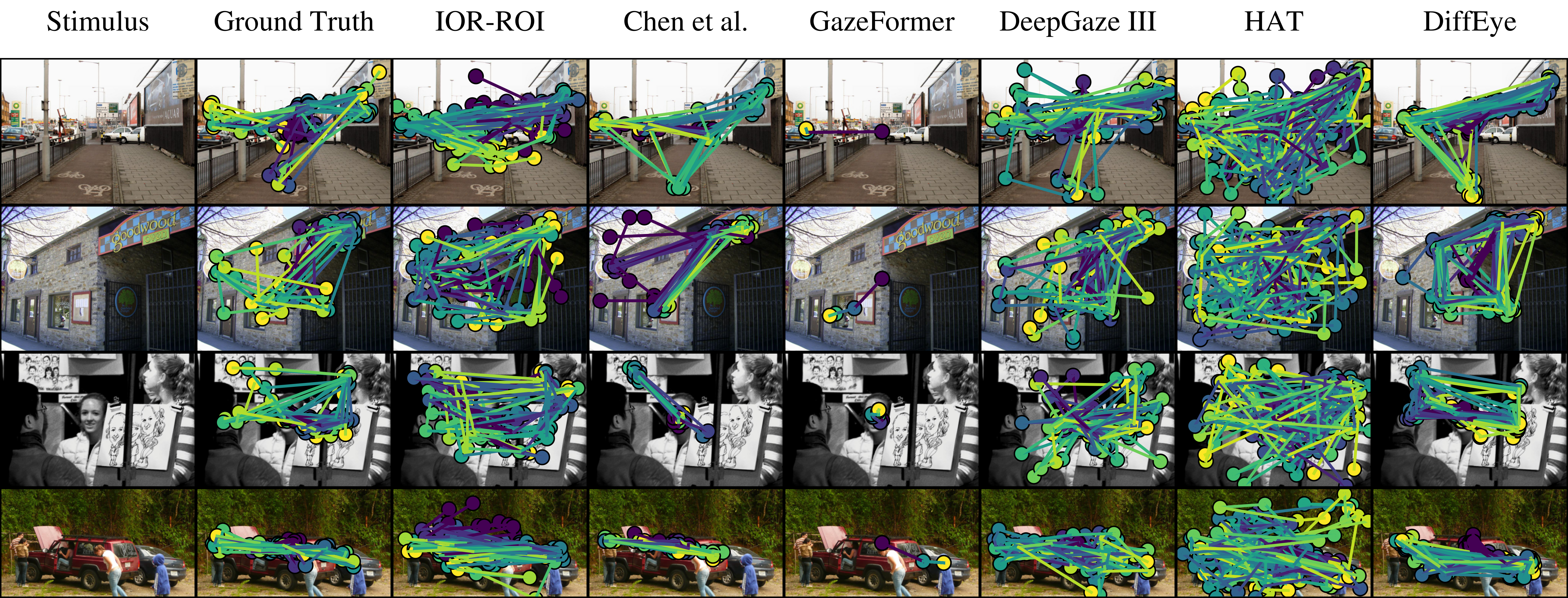}
    \caption{\textbf{\textit{Additional scanpath generation results.}} 
Scanpaths generated by \texttt{DiffEye} and baseline models are shown alongside ground truth annotations across four different scenes. Each row represents a unique stimulus, and each column shows the generated scanpaths for each method.}
    \label{fig:a3}
\end{figure}

Please see additional examples of scanpaths generated by \texttt{DiffEye} for the MIT1003 dataset in Fig.~\ref{fig:a3}.

\subsection{Additional Continuous Eye Movement Trajectory Distributions} \label{a:cont}
\begin{figure}[H]
    \centering
    \includegraphics[width=.8\linewidth]{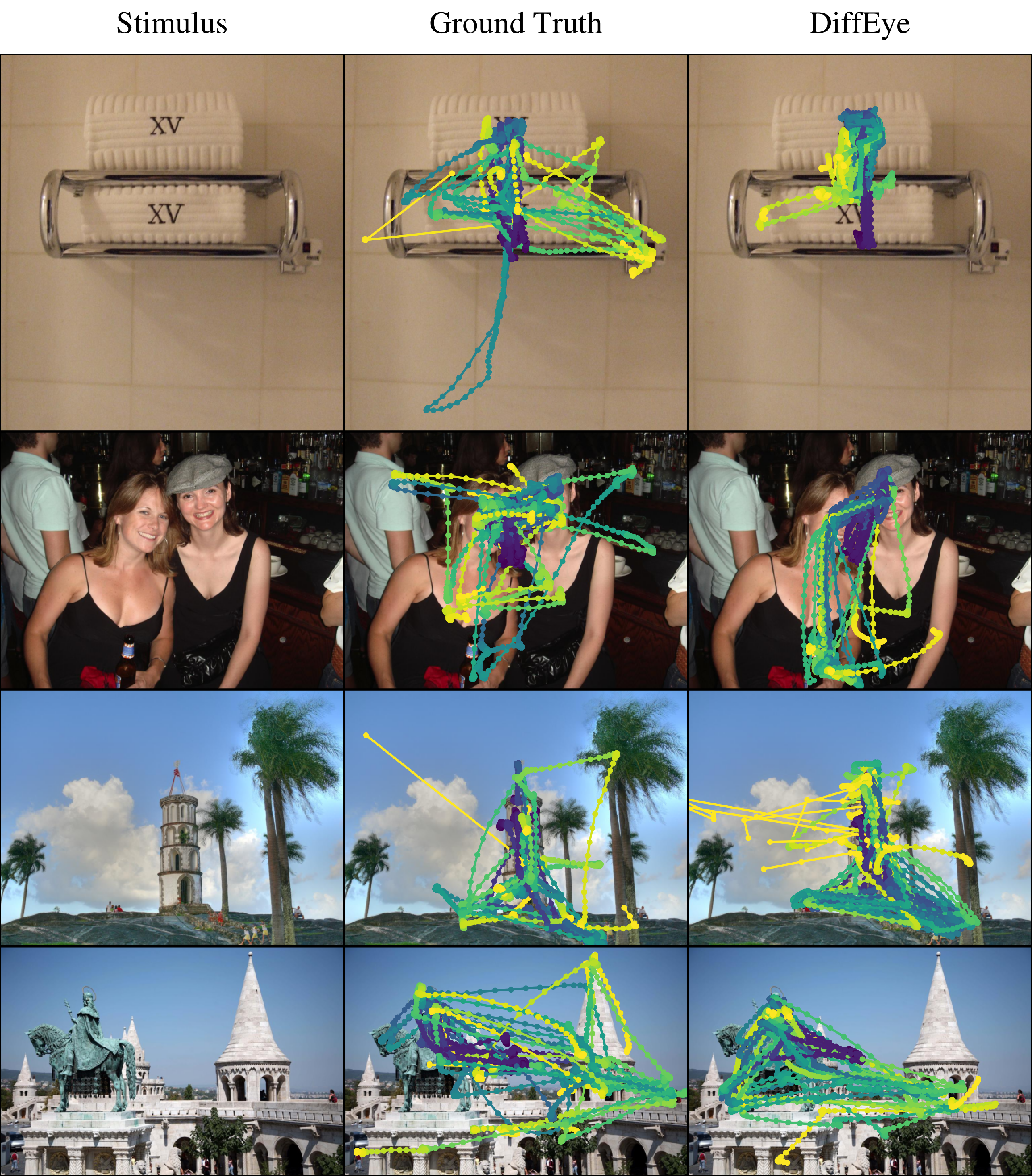}
    \caption{\textbf{\textit{Additional qualitative results of continuous eye movement trajectory generation.}} 
Additional eye movement trajectories generated by \texttt{DiffEye} alongside ground truth annotations across four different scenes. }
    \label{fig:a4}
\end{figure}

Please see additional examples of continuous eye movement trajectories generated by \texttt{DiffEye} for the MIT1003 dataset in Fig.~\ref{fig:a4}.

\subsection{Analysis of Saliency Prediction} \label{a:sal_analysis}
\noindent \paragraph{Saliency Prediction} To evaluate the spatial realism of our generated eye-tracking sequences, we compared our method against several models trained specifically for the saliency prediction task. The baselines include: the SUM model \cite{Hosseini2025sum}, which integrates the Mamba architecture with a U-Net to output saliency maps across diverse image types; TranSalNet \cite{Lou2022transalnet}, which leverages transformers for saliency prediction; and DeepGaze I \cite{Kümmerer2014deepgazeI} and DeepGaze IIE \cite{Linardos2021}, which are based on pretrained convolutional neural networks. 

\begin{figure}[H]
    \centering
    \includegraphics[width=.8\linewidth]{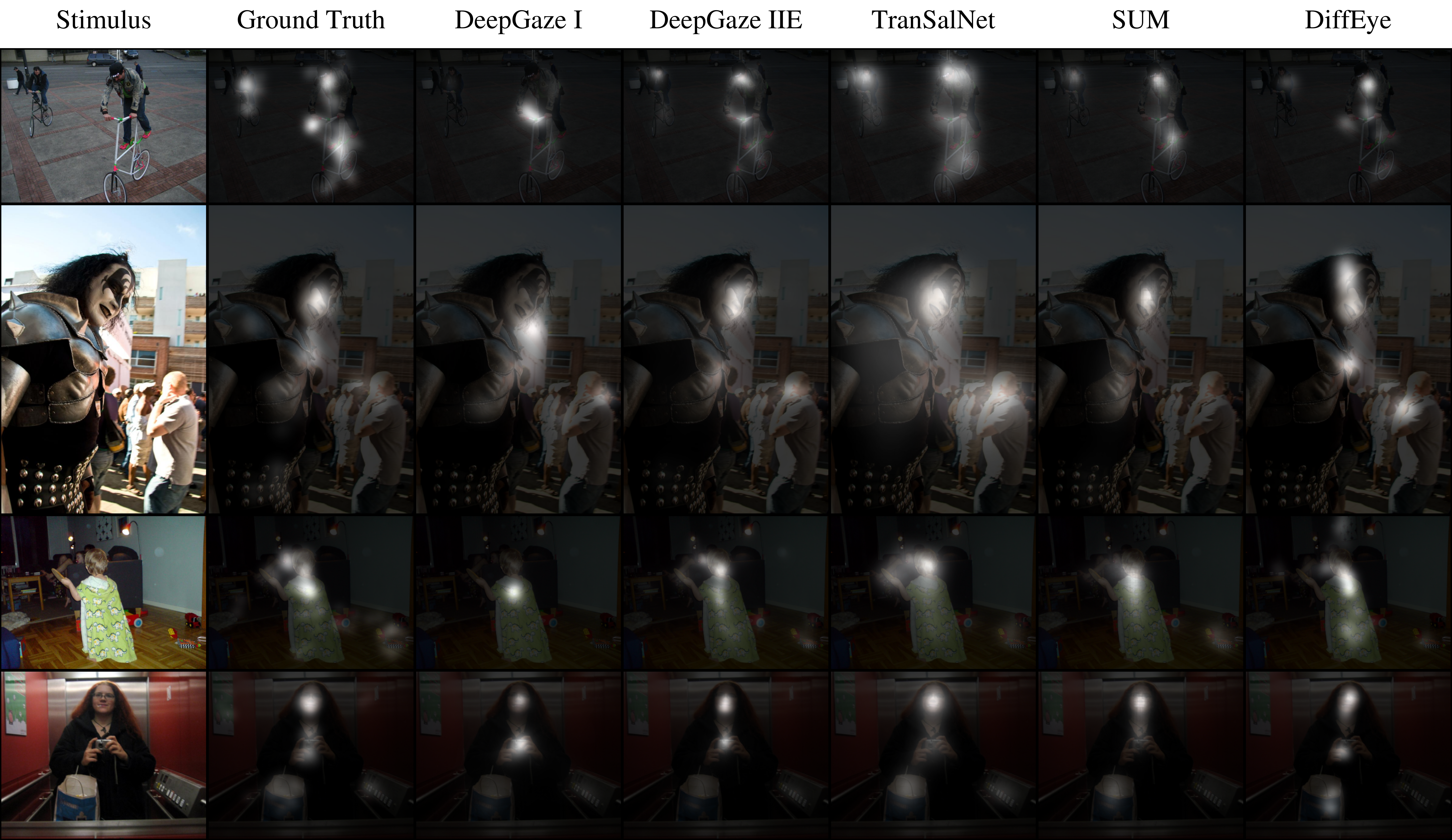}
    \caption{\textbf{\textit{Additional qualitative results for saliency prediction.}} 
Saliency maps generated by \texttt{DiffEye} and baseline models are shown alongside ground truth maps for four different scenes. 
Each row corresponds to a different stimulus image, with columns displaying the stimulus, ground truth saliency map, and predictions.}
    \label{fig:a5}
\end{figure}

 For each image in the test set, we generated 15 eye-tracking trajectories using our method. From each trajectory, we extracted fixation points using a script provided by \cite{Judd2009}, and used these to create individual fixation maps. These maps were aggregated and convolved with a Gaussian kernel to produce a single saliency map per image. For the baseline methods, which directly output saliency maps, we passed the same test images through each model. We then compared all predicted saliency maps, including ours, to the ground truth saliency maps provided in the dataset using six standard metrics: AUC-Judd, AUC-Borji, Normalized Scanpath Saliency (NSS), Similarity (SIM), Pearson's Correlation Coefficient (CC), and Kullback–Leibler Divergence (KL). Please refer to \cite{Bylinskii2019} for a comprehensive detailing of the saliency metrics used. Fig.~\ref{fig:a5} shows additional examples of saliency maps generated by \texttt{DiffEye} and the baselines for the MIT1003 dataset and Table~\ref{saliency-quantitative} reports the quantitative results.

\begin{table}[H]
  \caption{Saliency prediction comparison. Bold is best and underline is second best.}
  \label{saliency-quantitative}
  \centering
  \resizebox{0.7\linewidth}{!}{%
  \begin{tabular}{lcccccc}
    \toprule
    \textbf{Method}       & \textbf{AUC-Judd} $\uparrow$ & \textbf{AUC-Borji} $\uparrow$ & \textbf{NSS} $\uparrow$ & \textbf{SIM} $\uparrow$ & \textbf{CC} $\uparrow$ & \textbf{KL} $\downarrow$ \\
    \midrule
    DeepGaze I            & 0.883                        & 0.766                         & 2.306                   & 0.484                   & 0.580                  & \underline{0.980}         \\
    DeepGaze IIE          & \underline{0.923}            & 0.830                         & \underline{3.321}       & \underline{0.618}       & \underline{0.794}      & \textbf{0.552}            \\
    TranSalNet            & 0.896                        & \textbf{0.874}                & 2.443                   & 0.508                   & 0.658                  & 6.369                     \\
    SUM                   & \textbf{0.931}               & \underline{0.8458}            & \textbf{3.611}          & \textbf{0.727}          & \textbf{0.878}         & 1.438                     \\
    \texttt{DiffEye}      & 0.832                        & 0.737                         & 1.991                   & 0.447                   & 0.527                  & 1.991                     \\
    \bottomrule
  \end{tabular}
  }
\end{table}

\newpage
\subsection{Analysis of Statistical Properties of Scanpath Generation} \label{a:stats_scanpath}

\begin{figure}[!ht]
    \centering 
    \begin{subfigure}[b]{0.48\textwidth}
        \centering
        \includegraphics[width=\textwidth]{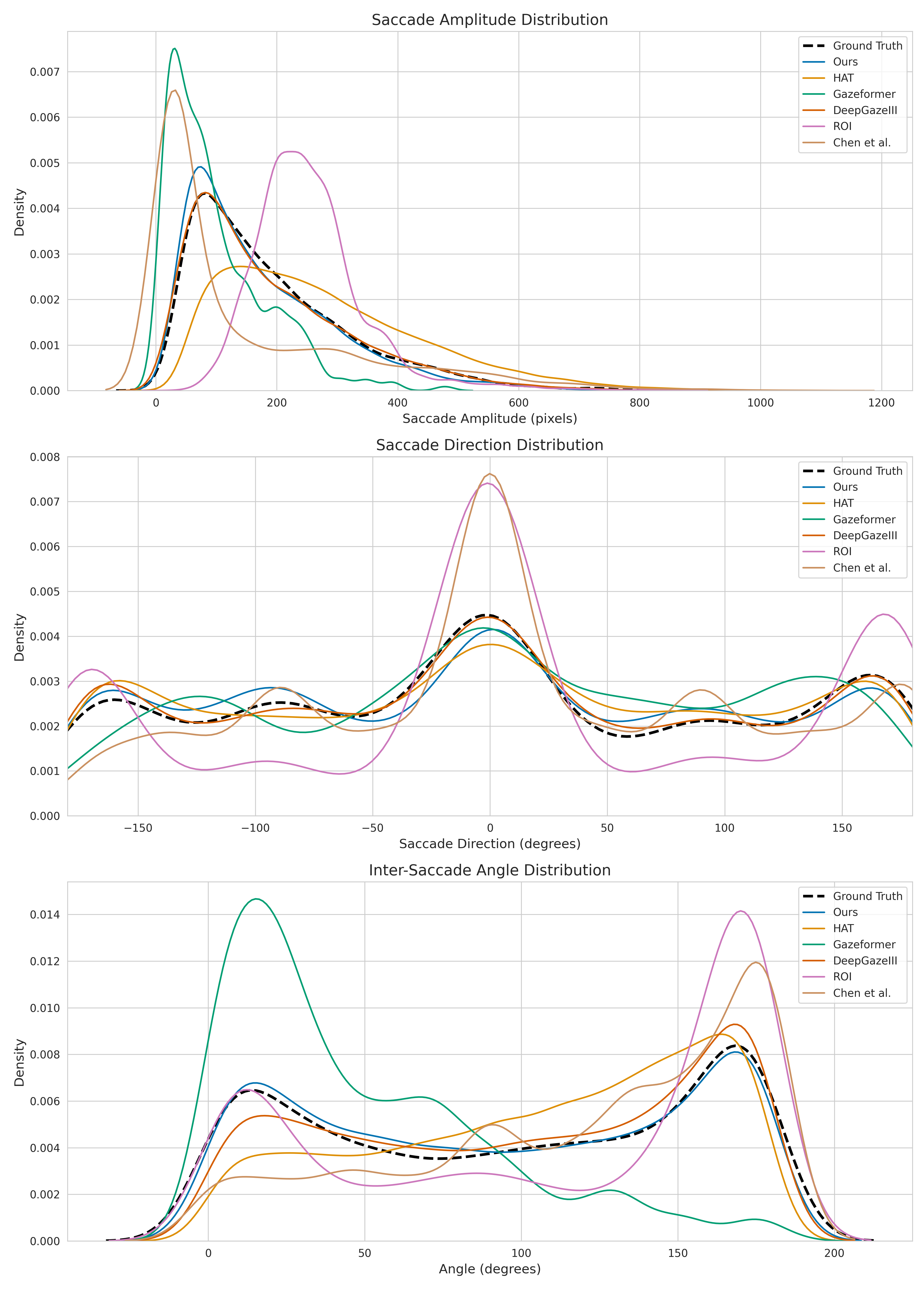}
        \caption{MIT1003}
        \label{fig:stat-mit}
    \end{subfigure}
    \hfill 
    \begin{subfigure}[b]{0.48\textwidth}
        \centering
        \includegraphics[width=\textwidth]{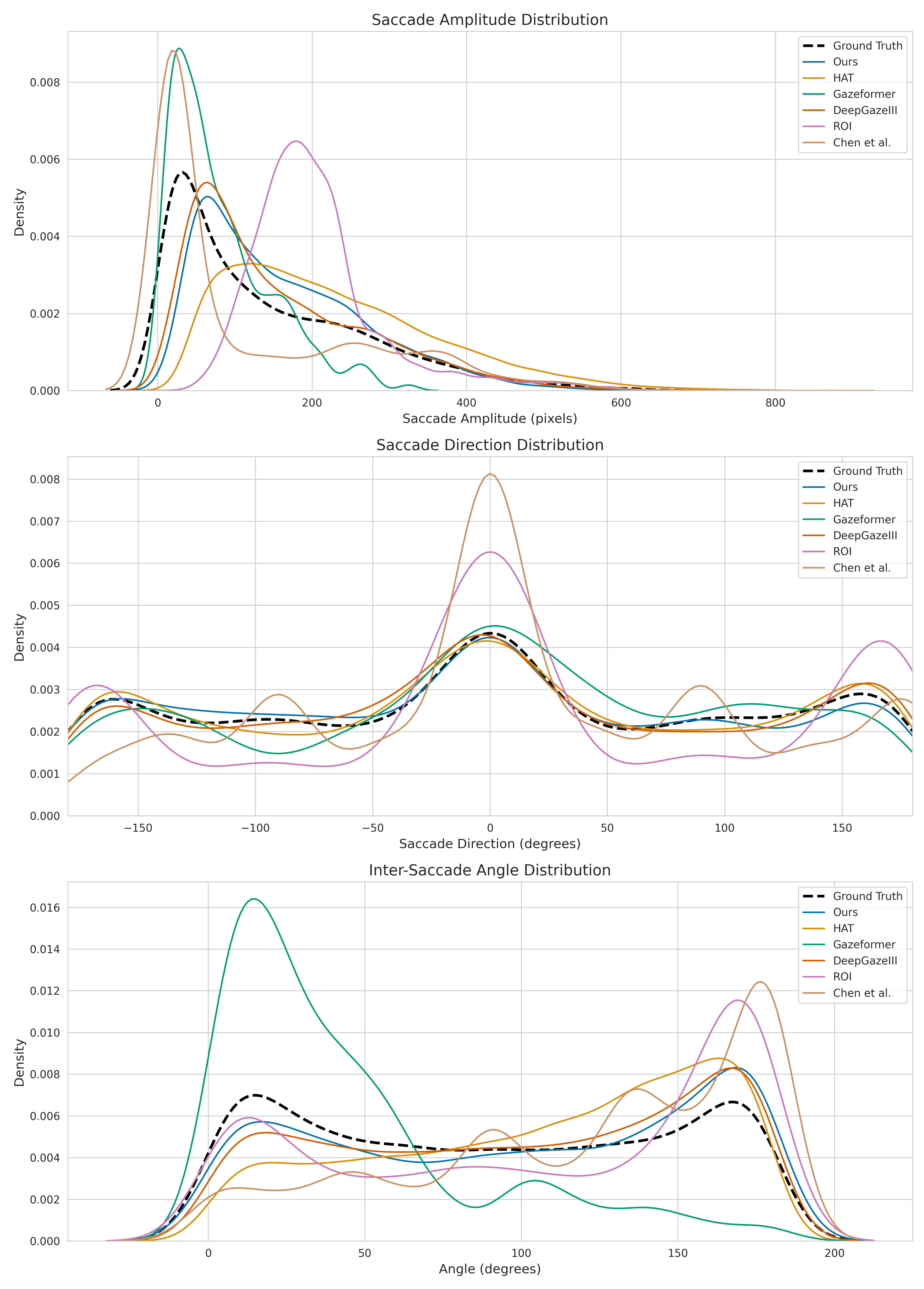}
        \caption{OSIE}
        \label{fig:stat-osie}
    \end{subfigure}
    \caption{Comparison of statistical properties for generated scanpaths on the (a) MIT1003 and (b) OSIE datasets. The distributions for saccade amplitude, saccade direction, and inter-saccade angle of our model (blue) are compared against ground truth human scanpaths (black, dashed) and several other methods. Our model consistently provides the closest fit to the ground truth distributions.}
    \label{fig:stats_scanpath}
\end{figure}

To evaluate the plausibility of the generated scanpaths, we compare their statistical properties against those of ground truth human eye movements. Figure~\ref{fig:stats_scanpath} shows the distributions of three key metrics: saccade amplitude, saccade direction, and inter-saccade angle for both the MIT1003 and OSIE datasets.

Our model's performance (blue line) demonstrates a strong alignment with the ground truth distributions (black dashed line) across all three metrics. In the saccade amplitude distribution, our model successfully captures the peak at lower pixel values. For saccade direction, our model accurately reflects the horizontal bias present in human vision (peaks at 0 and $\pm180$ degrees). Finally, in the inter-saccade angle distribution, our model correctly shows a strong tendency for forward movements (peak near 0 degrees) and return saccades (smaller peak near $\pm180$ degrees). These results are consistent across both datasets, confirming that our approach generates statistically more realistic scanpaths than the compared methods.

\end{document}